\documentclass[runningheads]{llncs}

\usepackage[mobile]{eccv}

\usepackage{eccvabbrv}

\usepackage{graphicx}
\usepackage{booktabs}
\usepackage{marvosym}
\usepackage[accsupp]{axessibility}  

\usepackage{hyperref}

\usepackage{orcidlink}

\makeatletter
\def\blfootnote{\gdef\@thefnmark{}\@footnotetext}
\makeatother

\begin{document}

\title{DDStereo: Efficient Dual Decoder Transformers for Stereo 3D Road Anomaly Detection} 

\titlerunning{Dual Decoder Transformers for 3D Anomaly Detection}

\author{Shiyi Mu\inst{1}\orcidlink{0000-0002-5393-0258} \and
Zichong Gu\inst{1}\orcidlink{0009-0007-5787-252X} \and
Zhiqi Ai\inst{1}\orcidlink{0009-0005-1034-9972} \and
Yilin Gao\inst{1}\orcidlink{0000-0003-4985-4034}\and
Shugong Xu\inst{2}\orcidlink{0000-0003-1905-6269}\textsuperscript{\Letter}} 

\authorrunning{S.~Mu et al.}

\institute{Shanghai University, Shanghai, China \\
\and
Xi’an Jiaotong-Liverpool University, Suzhou, China \\
\email{shugong.xu@xjtlu.edu.cn}
}
\maketitle

\blfootnote{\textsuperscript{\Letter} Shugong Xu is the  corresponding author.}

\begin{figure}[t]
  \centering
  \begin{subfigure}{0.5\linewidth}
    \centering
    \includegraphics[width=\linewidth]{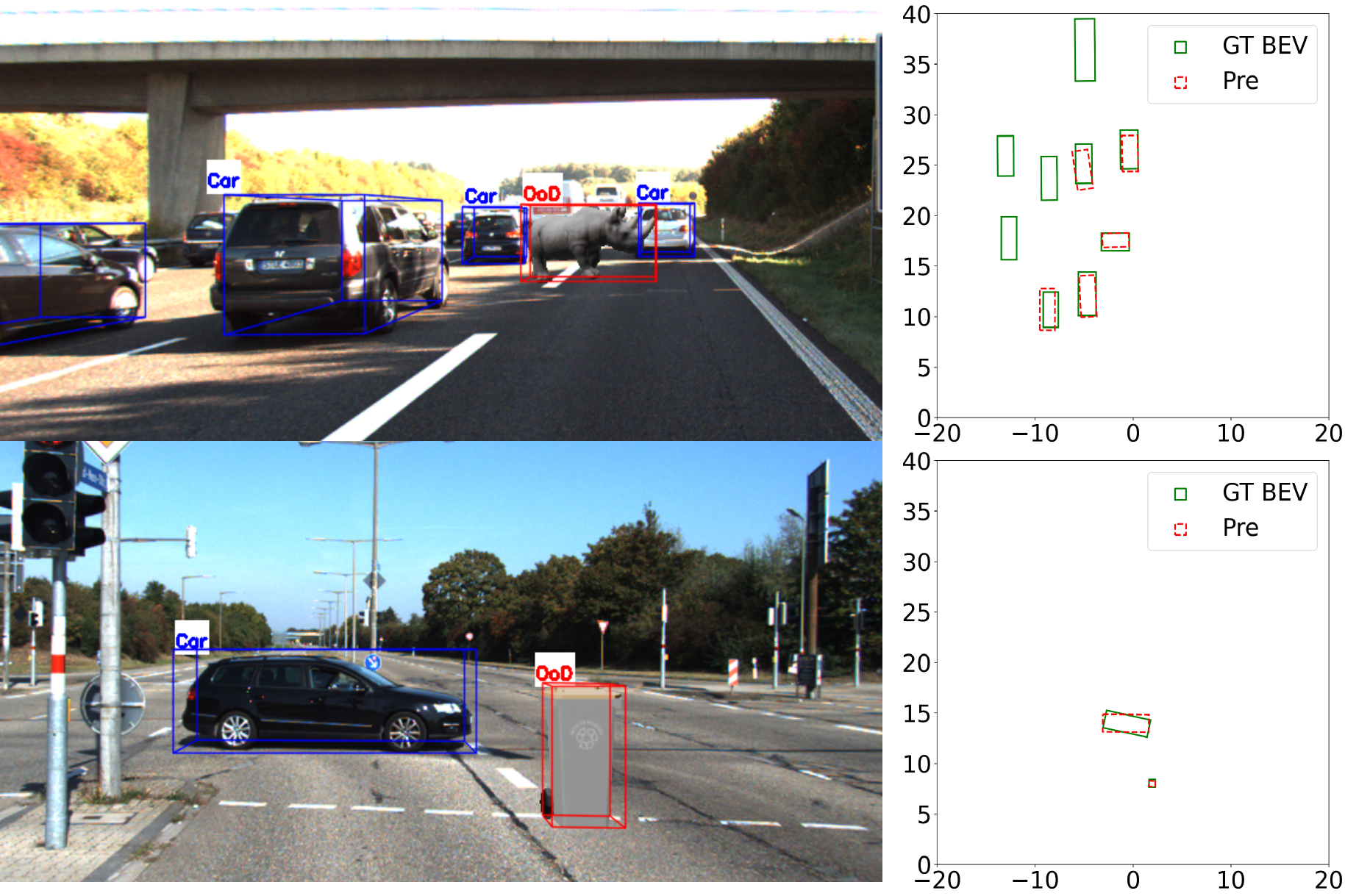}
    \caption{Visualization of 3D road anomaly detection}
    \label{fig:visual_anomaly}
  \end{subfigure}
  \hfill
  \begin{subfigure}{0.45\linewidth}
    \centering
    \includegraphics[width=\linewidth]{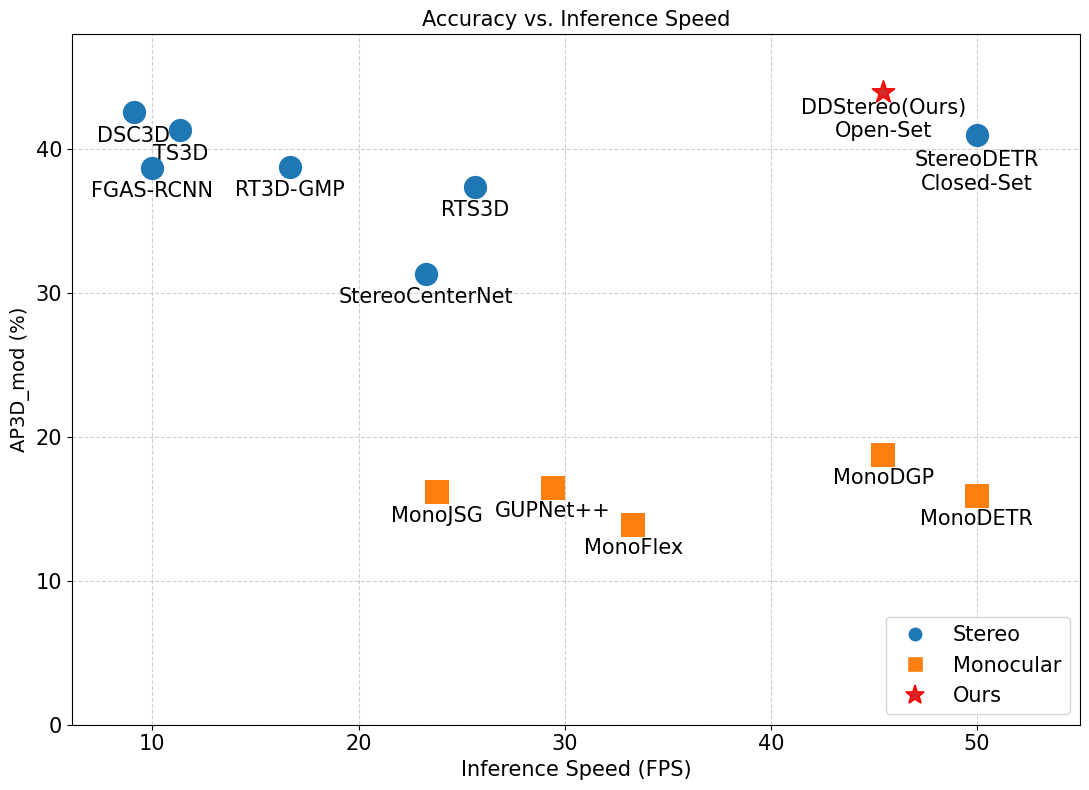}
    \caption{Speed vs. Accuracy}
    \label{fig:speed_acc_comparison}
  \end{subfigure}

  \caption{
    \textbf{Qualitative and quantitative results.} 
    (a) Our model demonstrates robust 3D road anomaly detection capabilities on AR-OoD\cite{S3AD_mu25}, providing precise 3D bounding boxes and BEV representations. 
    (b) Performance comparison with closed-set monocular and stereo 3D detection methods on KITTI\cite{KITTI}. Our approach achieves a superior trade-off between inference speed, accuracy and open-set.
  }
  \label{fig:main_results}
\end{figure}

\begin{abstract}
  
Stereo-based 3D obstacle perception for autonomous driving is currently constrained by an imbalanced triplet: deployment cost, detection accuracy, and open-set adaptability. While existing methods struggle to balance these three competing objectives, there is an urgent demand for high-precision, real-time algorithms capable of detecting arbitrary obstacles in the wild. In this paper, we present DDStereo, a novel Dual-Decoder Stereo Transformer that achieves a synergistic integration of 3D object detection and Out-of-Distribution (OoD) road anomaly detection. Leveraging the geometric priors of stereo disparity, our approach effectively couples 3D attribute regression with open-set foreground detection within a streamlined dual-branch decoder architecture. Conventional methods rely on complex feature-level fusion; DDStereo maintains execution efficiency by employing a decoupled decoding strategy and shared object-level queries to ensure cross-modal target alignment. Extensive evaluations of public benchmarks demonstrate that DDStereo not only achieves state-of-the-art accuracy under open-set and closed-set protocols. Our method delivers real-time performance comparable to monocular 3D detection baselines, providing a cost-effective solution for the perception of obstacles of the normal and OoD category. Code and models are available at https://github.com/shiyi-mu/DDStereo.
\end{abstract}

\section{Introduction}
\label{sec:intro}

\begin{quote}
\textit{``To sense the unseen is the threshold of safety.''} 
\end{quote}

In the context of intelligent driving, the ``unseen'' refers to the vast array of unforeseen, Out-of-Distribution (OoD) obstacles that lie beyond fixed training labels. Safely navigating these anomalies has long been the primary bottleneck for stereo-based systems, which must reconcile high-fidelity geometric reasoning with the low-latency demands of real-time deployment.

3D object detection has become a cornerstone technology in autonomous driving and robotic perception. While LiDAR-based methods dominate in accuracy, stereo vision offers a cost-effective alternative by providing dense depth without expensive sensors. However, a significant gap remains: most stereo 3D detectors have reduced latency to around 40\, ms, yet they remain nearly twice as slow as their monocular counterparts. StereoDETR\cite{StereoDETR} achieves a speed comparable to monocular detection, but it remains limited to closed-set detection.
Simultaneously, while monocular detection has advanced toward open-vocabulary recognition \cite{ovmono3d_arx24, DetAny3D_iccv25}, most stereo approaches still operate under a \textbf{closed-set assumption}, rendering them blind to unseen categories such as road debris or novel construction equipment. In real-world environments, this assumption is frequently violated. Autonomous agents must not only localize known objects but also respond to OoD instances. Recent efforts have introduced stereo-based road anomaly detection; however, their inference speed exceeds 80\,ms\cite{S3AD_mu25}, far from the requirements of real-time deployment. Furthermore, unlike existing open-vocabulary algorithms that require heavy text prompts or manual cues, the immediate demand in driving scenarios is prompt-free, open-ended detection. Rather than predicting fine-grained category names, the priority is the rapid binary classification of an obstacle's presence to ensure immediate safety. To address these challenges, we redefine stereo-based 3D perception as a unified optimization of a fundamental triplet: accurate multi-class detection, robust binary OoD discovery, and ultra-fast real-time inference. 
Fig.~\ref{fig:main_results} illustrates the detection results and the balance between speed and accuracy.
\begin{figure}[h]
\centering
\includegraphics[height=3.5cm]{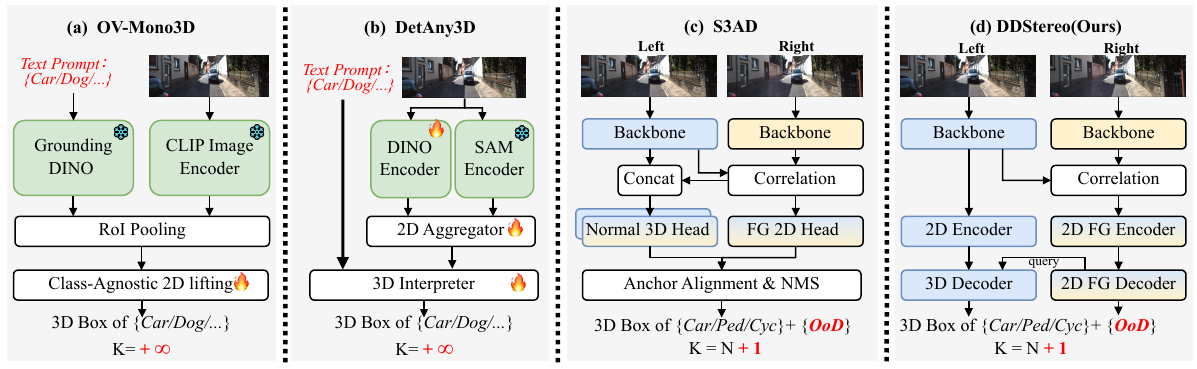}
\caption{Compare with other methods. OV-Mono3D\cite{ovmono3d_arx24} and DetAny3D\cite{DetAny3D_iccv25} rely on text prompt matching, whereas S3AD\cite{S3AD_mu25} uses anchor-level fusion to detect unknown foregrounds; neither approach is real-time. Our DDStereo decouples texture and disparity features to independently predict 3D attributes and foreground objectness, maintaining high computational efficiency.
}
\label{fig:campare}
\end{figure}

The risk of missed detections is significantly exacerbated for unknown categories. While a small fraction of novel objects may be misclassified into known classes, the vast majority are typically overlooked due to the absence of category-specific priors and supervision. Existing open-vocabulary detectors attempt to mitigate this by requiring external guidance, such as textual prompts \cite{ovmono3d_arx24} or interactive cues (e.g., points or boxes \cite{DetAny3D_iccv25}). However, in high-stakes autonomous driving, a truly robust paradigm must rely solely on visual input to perceive all potential obstacles, regardless of their semantic labels. As shown in Fig.~\ref{fig:campare}, while methods like OV-Mono3D \cite{ovmono3d_arx24} and DetAny3D \cite{DetAny3D_iccv25} enable unconstrained recognition via language-vision alignment, they introduce significant computational overhead. Conversely, S3AD \cite{S3AD_mu25} extends $N$-class detection by treating OoD objects as an auxiliary $(N{+}1)$-th class. Despite its effectiveness, S3AD suffers from prohibitive latency ($>$80\,ms) due to dense anchor-based Non-Maximum Suppression (NMS) and high-dimensional feature concatenation.

To address these challenges, we propose \textbf{DDStereo}, a Dual-Decoder Stereo Transformer designed for end-to-end open-set 3D detection. Our core philosophy is the \textbf{\textit{decoupling of category-agnostic localization from semantic classification}}. Specifically, a standard decoder regresses 3D attributes for In-Distribution (ID) objects, while a specialized foreground decoder identifies potential instances leveraging disparity-aware features. Since disparity is fundamentally driven by geometric consistency rather than texture patterns, this branch enables robust localization of novel objects based on their physical presence alone. 

To ensure spatial alignment between these decoupled branches, we introduce a \textbf{shared object query} mechanism. This design maintains strict consistency between foreground boxes and class-specific predictions, allowing anomaly scores to be derived from the discrepancy between foreground confidence and semantic classification probability. Furthermore, to avoid memory-intensive feature concatenation, we decouple depth estimation from attribute regression. A lightweight convolutional head predicts object-wise depth maps, which are efficiently fused via center-based sampling. This architecture not only eliminates the need for NMS but also slashes inference latency to 24\,ms.

The contributions of this work are summarized as follows:
\begin{itemize}
    \item \textbf{Unified Real-Time Open-Set Framework:} We present DDStereo, the first 3D detector to integrate high-precision, real-time inference and open-set adaptability into a single stereo pipeline.
    \item \textbf{Efficient Query-Base Alignment:} We introduce a shared object query mechanism that elegantly bridges 3D regression and anomaly detection. This simple yet efficient design ensures cross-task consistency while eliminating the need for complex feature-level fusion.
    \item \textbf{Generalization to the Wild:} We validate thegeneralization potential of DDStereo through extensive evaluations on out-of-distribution categories and diverse real-world visualizations, demonstrating robust obstacle perception in unconstrained environments.
\end{itemize}

\section{Related Work}

\subsection{Stereo 3D Object Detection}
Stereo-based 3D object detection methods can be broadly categorized by how they compute disparity or depth cues, typically falling into three paradigms: image-level disparity, region-level disparity, and feature-level disparity. Early approaches rely on image-level disparity estimation, such as RT3DStereo~\cite{RT3DStereo_2019} and RT3D-GMP~\cite{RT3D-GMP-ITSC20}. However, such pipelines often suffer from depth noise and are computationally intensive. Region-level disparity methods extract object-aligned regions using 2D detectors and compute disparity only within those areas as TLNet~\cite{TLNet_CVPR19}, Stereo R-CNN~\cite{Stereo-R-CNN-2019}, and SIDE~\cite{SIDE_WACV22}.
Recent works favor feature-level disparity learning, which leverages stereo feature matching to directly regress 3D attributes from cost volumes or concatenated features\cite{YOLOStereo3D_2021, TS3D_TITS24, DSC3D_TCSVT25}. 
Although these methods have shown strong accuracy. They typically rely on heavy anchor based heads or 3D cost volumes, leading to limited runtime performance. With an extremely lightweight design, StereoDETR\cite{StereoDETR} breaks the speed bottleneck of monocular detection for the first time, yet it is restricted to closed-set scenarios.

\subsection{Open-Set and Unknown 3D Object Detection}
Open-set object detection (OSOD) and open-vocabulary object detection (OVOD) have been explored as two complementary paradigms to detect unknown objects. In 2D field, OVOD methods learn visual text alignments to localize arbitrary categories specified by text queries 
\cite{yao2023detclipv2,gu2021open,wu2023cora,cheng2024yolo,liu2023grounding,chuang2024generative,yao2024detclipv3},
while OSOD approaches aim to discover any objects absent from the training set, typically labeling them as “unknown” \cite{dhamija2020overlooked,han2022expanding,gupta2022ow,joseph2021towards,yao2022detclip}.
In 3D field, existing work has focused almost exclusively on OVOD, leveraging 2D open-vocabulary detectors to lift category-agnostic proposals into class-agnostic 3D detectors \cite{cen2021open,lu2023open,cao2024coda,wang2024ov,xue2023ulip,zhang2022pointclip}. Open-set 3D detection, however, has received limited attention. Cen et al. \cite{cen2021open} first formalize the task and propose a two-stage pipeline that scores point-wise entropy to reject unseen classes. Alliegro et al. \cite{20223DOS} introduce 3DOS, a benchmark and baseline that exposes the sensitivity of point-cloud detectors to semantic novelty. Lu et al. \cite{lu2023open} adopt 2D open-vocabulary knowledge to generate pseudo-labels for unknown objects yet stop short of explicit open-set modeling. 
S3AD~\cite{S3AD_mu25} is the first method to introduce open-set 3D object detection based on stereo vision, along with the release of the AR-OoD benchmark, aiming to detect arbitrary potential obstacles and enhance autonomous driving safety. However, as S3AD is built upon the YOLOStereo3D~\cite{YOLOStereo3D_2021} framework, its inference time exceeds 80\,ms, making it difficult to meet real-time requirements.

\section{Method}

\subsection{Overview}

\begin{figure}[t!]
\centering
\includegraphics[height=5.5cm]{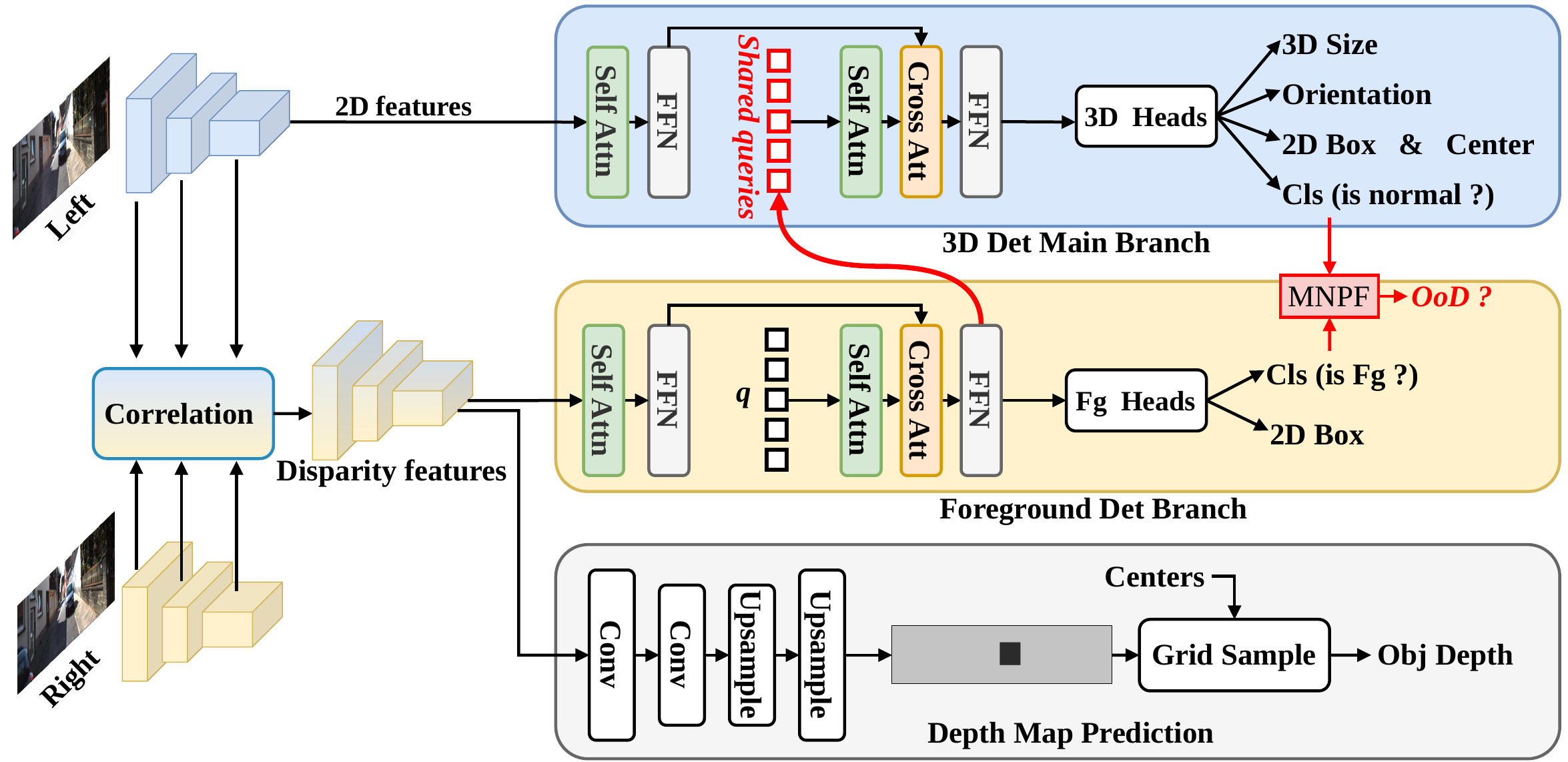}
\caption{Overview of the proposed DDStereo architecture. (1) The main branch predicts 3D attributes via monocular features; (2) the foreground branch outputs texture-agnostic binary foregrounds; and (3) the depth map branch estimates object-level center depth. All three branches are coupled using shared queries and Grid Sample.
}
\label{fig:framework}
\end{figure}

We propose DDStereo, a Dual-Decoder stereo 3D detection framework tailored for open-world scenarios. As shown in Fig.~\ref{fig:framework}, our model consists of three collaborative branches: a main 3D detection branch, a foreground detection branch, and an object-level depth map prediction branch. These branches are tightly integrated through query-level and point-level alignment strategies. 

Let $\mathcal{I}_L$ and $\mathcal{I}_R$ denote the left and right stereo images, respectively. We extract 2D visual features from the left image only using a Resnet backbone, denoted as $f_{V}$. Meanwhile, to encode stereo feature, we compute a correlation volume from the features of $\mathcal{I}_L$ and $\mathcal{I}_R$ via cross-view cost aggregation. The resulting disparity-based feature map, denoted as $f_{D}$, encodes stereo-aware depth information.

The 3D detection main branch operates on $f_{V}$ and utilizes a transformer-based encoder-decoder architecture to predict 3D object properties, including size, orientation, class scores (for known categories), and 2D bounding boxes. Concurrently, the foreground detection branch utilizes $f_{D}$ to predict coarse 2D boxes and foreground probabilities via binary classification.
To align the 2D texture representations and stereo-derived depth cues, both detection branches are conditioned on a shared object queries, facilitating joint reasoning across appearance and depth.

To enhance depth map estimation, we upsample the stereo disparity feature $f_{D}$ to produce an object-level depth map, denoted as $\mathbf{M}_{\text{obj}} $. Given the centers of predicted boxes from the main 3D detection branch, we use a grid sampling operation to extract object-centric depth values, which are further used to complement 3D box localization. Finally, we introduce a MNPF module, where MNPF refers to the comparison between the maximum normal probability and the foreground confidence, to estimate the open-set confidence of each detection. This enables our model to identify OoD objects beyond the training categories.

\subsection{Visual feature and Disparity feature}
The left and right stereo images are first processed through a shared backbone to extract multi-scale features at \(\frac{1}{4}\), \(\frac{1}{8}\), \(\frac{1}{16}\) , and \(\frac{1}{32}\) resolutions. The multi-scale features from the left view are used as inputs to the visual decoding branch.

\begin{figure}[t]
  \centering
  \begin{subfigure}{0.38\linewidth}
    \centering
    \includegraphics[width=\linewidth]{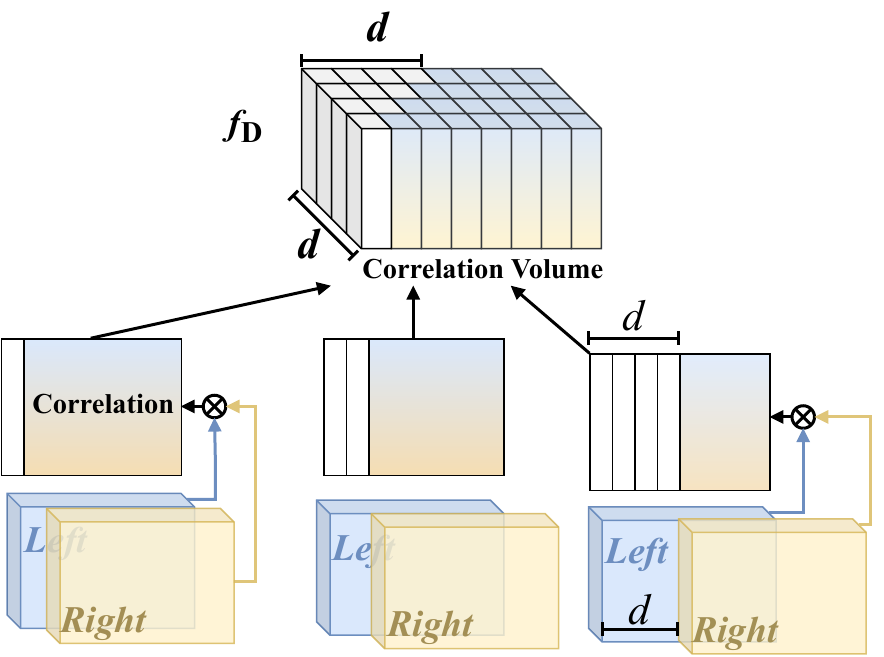}
    \caption{Stereo correlation volume}
    \label{fig:stereo_cost}
  \end{subfigure}
  \hfill
  \begin{subfigure}{0.58\linewidth}
    \centering
    \includegraphics[width=\linewidth]{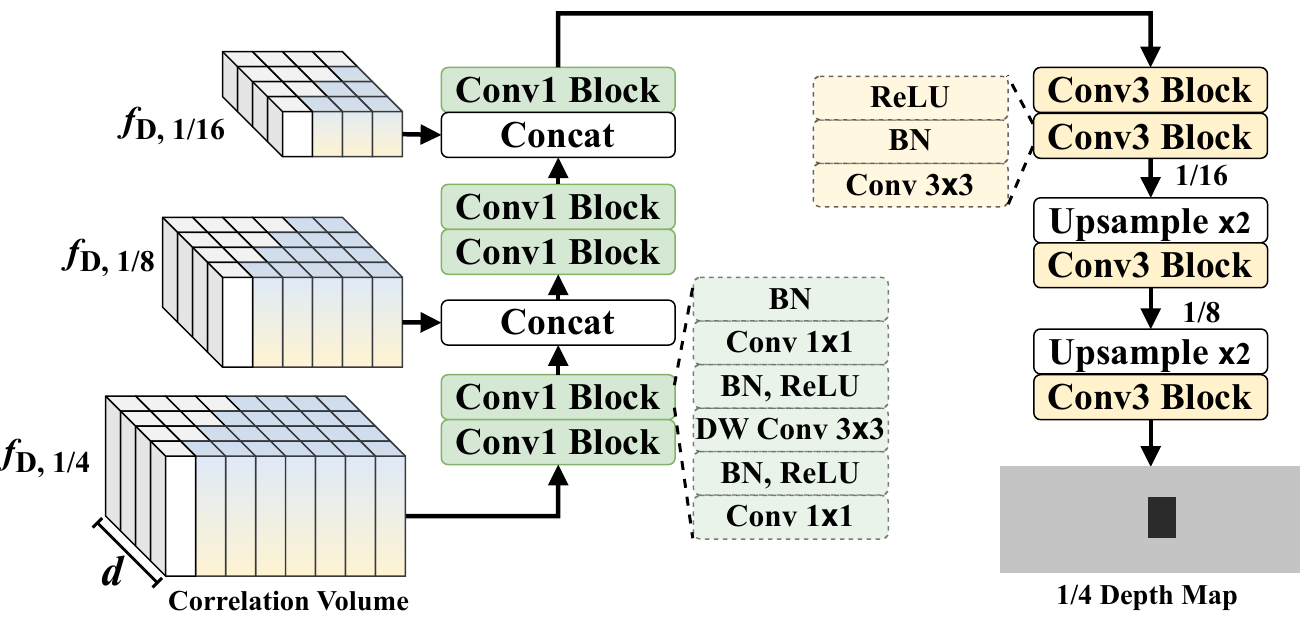}
    \caption{Object level depth prediction}
    \label{fig:depth_prediction}
  \end{subfigure}

  \caption{
    \textbf{Overview of the stereo matching and depth estimation modules.} 
    (a) The computation process of the stereo correlation volume, constructed from concatenated left-right feature maps. 
    (b) Our multi-scale disparity fusion strategy for high-precision depth map prediction, providing the geometric foundation for subsequent 3D detection.
  }
  \label{fig:stereo_depth_modules}
\end{figure}

To construct disparity-aware representations, we compute simple correlation volumes between the left and right features at each scale. As illustrated in Fig.~\ref{fig:stereo_cost}, we generate these disparity features by shifting the right-view features along the horizontal axis up to a maximum displacement of \(d\) pixels and calculating per-pixel correlations with the left-view features. 
The maximum disparity steps for the four scales are set to 24, 24, 16, and 16.

\subsection{Depth map decoder and Sampling}

In MonoDETR\cite{monodetr_iccv23} and MonoDGP\cite{Monodgp_cvpr25}, an explicit and coarse-grained \( \frac{1}{16} \) depth map is predicted, where the ground-truth is constructed from the 3D center distance and the corresponding 2D bounding box of each object. The predicted depth is further encoded and fused via a depth encoder and cross-attention to guide the 3D decoding process. Similar to the architecture of StereoDETR\cite{StereoDETR}, we adopt a more efficient strategy that predicts a higher-resolution depth map and directly samples depth values at the predicted object centers from the detection branch. As shown in Fig.~\ref{fig:depth_prediction}, the depth prediction module adopts a simple U-shaped architecture. Multi-scale correlation volumes are concatenated after down-sampling and then upsampled through convolutional layers to reconstruct a depth map at \( \frac{1}{4} \) resolution. The projected 2D center of each 3D object is used to sample the predicted depth as the object's depth estimate. The ground-truth depth map and supervision loss follow the same setting as in MonoDETR\cite{monodetr_iccv23} and MonoDGP\cite{Monodgp_cvpr25}.

\subsection{Binary Foreground Detection Branch}

It is difficult to determine the specific category of an object from a depth or disparity map. However, it is often sufficient to identify whether an obstacle is present. To this end, we design a foreground detection branch as a binary DETR-style decoder based on disparity features.
Specifically, the multi-scale disparity features \(f_D\) are first flattened and fed into an encoder, which consists of a single-layer multi-scale deformable attention module followed by a feed-forward network (FFN). The output embeddings are denoted as \(f^{e}_{\text{D}}\).

In the decoder, we define a learnable object query \(q \in \mathbf{R}^{N \times C}\), where \(N\) is the maximum number of objects. This same query is also shared with the 3D detection decoder. The decoder is composed of an inter-query self-attention layer, a visual cross-attention layer, and a feed-forward network. Finally, two linear layers are used to predict the foreground classification and the 2D bounding boxes.
\begin{align}
\text{SelfAttn}(q) &= \text{softmax}\left( \frac{q q^\top}{\sqrt{C}} \right) q, \\
\text{CrossAttn}(q, f^{e}_{\text{D}}) &= \text{softmax}\left( \frac{q (f^{e}_{\text{D}})^\top}{\sqrt{C}} \right) f^{e}_{\text{D}}.
\end{align}

\subsection{3D Detection Branch}
Unlike MonoDETR\cite{monodetr_iccv23} and MonoDGP\cite{Monodgp_cvpr25}, we do not adopt the depth-guided cross-attention mechanism in the 3D detection branch. Instead, we follow a structure similar to the original 2D detector to predict all attributes except depth, as the dual-decoder design already encodes geometry via shared queries, making explicit depth attention redundant and preserving computational efficiency.

Specifically, the left-view visual features \(f_V\) are fed into an encoder composed of a single-layer self-attention module followed by a FFN, producing embeddings denoted as \(f^{e}_{V}\). In the decoder, we reuse the object query \(q\) from the foreground decoder. The decoder consists of an inter-query self-attention layer, a visual cross-attention layer, and a FFN, followed by several linear layers that respectively predict the 3D object size, orientation, normal category (with \(K\) classes), and the 2D bounding box. Notably, the 2D bounding box is parameterized using the projected 3D object center on the image plane as the reference point. The box is represented by offsets from the center to the top, bottom, left, and right boundaries, resulting in six parameters in total. This projected center will be used in the subsequent object depth sampling process.
\begin{align}
\text{CrossAttn}(q, f^{e}_{\text{V}}) &= \text{softmax}\left( \frac{q (f^{e}_{\text{V}})^\top}{\sqrt{C}} \right) f^{e}_{\text{V}}.
\end{align}

\subsection{Anomaly Confidence Estimation}
While previous road anomaly segmentation methods primarily focus on global pixel-level OoD scoring, we propose an object-level anomaly score estimation. Specifically, the anomaly score MNPF is computed as the distance between the \textbf{M}aximum \textbf{N}ormal \textbf{P}robability and the \textbf{F}oreground probability $S_f$:

\begin{equation}
S_{\text{OoD}} = S_f - \max_{i \in K} (\sigma(S_i)),
\end{equation}
where $K$ is the set of normal known classes. 
We set $\sigma$ as the sigmoid function. If replaced with a softmax operation, the latter term becomes equivalent to MSP~\cite{MSP_ICLR17}. If $\sigma$ is removed (no normalization is applied), it is equivalent to MaxLogit~\cite{MaxLogit_ICML22}.

\subsection{Loss Function}
The training loss consists of three main components: the 2D foreground loss, the 3D detection loss for in-distribution categories, and the disparity supervision loss. Both the 2D and 3D branches employ separate Hungarian matching processes for target assignment. The 2D loss includes a binary classification loss and a bounding box regression loss. The 3D prediction head regresses multiple attributes, including object category, orientation, 3D size, 3D center, and 2D bounding box.

\begin{equation}
\mathcal{L}_{\text{total}} = \lambda_{2D} \cdot \mathcal{L}_{2D} + \lambda_{3D} \cdot \mathcal{L}_{3D} + \lambda_{D} \cdot \mathcal{L}_{\text{map}}
\end{equation}

\noindent
Hungarian matching is independently applied for both $\mathcal{L}_{2D}$ and $\mathcal{L}_{3D}$ to associate predicted queries with ground-truth objects.

\section{Experiments}

\subsection{Experimental Setup}

\noindent \textbf{Anomaly Stereo 3D Dataset.}
AR-OoD\cite{S3AD_mu25} is the most category-rich dataset for 3D road anomaly detection. We follow the exact same data split as S3AD\cite{S3AD_mu25}. The dataset contains 4,038 training and 2,347 testing samples, encompassing 105 categories in total. The training set includes 47 categories to enhance the model's generalization ability, while the remaining 58 novel categories are used to evaluate road anomaly detection. During the testing phase, these 58 novel anomaly categories are mapped into a single "OoD" category.

\noindent \textbf{Close-set Stereo 3D Dataset.} Unlike Waymo\cite{Waymo_2020_CVPR} and nuScenes\cite{nuScenes_2020_CVPR}, KITTI\cite{KITTI} and Argoverse\cite{Argoverse_2019_CVPR} provide paired stereo images. On KITTI\cite{KITTI}, following the split in YOLOStereo3D\cite{YOLOStereo3D_2021}, we use 3,712 images for training and 3,769 for validation, submitting our unlabeled test set (7,518 samples) predictions to the official server. Most baseline methods only report Car metrics, we evaluate on the Car, Pedestrian, and Cyclist categories. For the more challenging Argoverse\cite{Argoverse_2019_CVPR} dataset, we adopt the format and split from StereoDistill\cite{StereoDistill}, using 13,122 training and 5,015 validation samples.

\noindent \textbf{Implementation Details. }
We train our DDStereo using four NVIDIA RTX A6000 GPUs with 48,GB of memory. For a fair evaluation of inference efficiency, all testing is performed on a single RTX 4090D GPU. We use the Adam optimizer with an initial learning rate of 0.0002, and apply a learning rate decay factor of 0.1 at the 125 and 165 epochs. The batch size is set to 16.

\subsection{Open-Set Evaluation on AR-OoD}

\begin{table}[h]
\centering
\caption{Anomaly evaluation on AR-OoD\cite{S3AD_mu25}. OoD refers to 58 novel categories mapped to a single class, while Pedestrian Car and Cyclist are known categories. Results are reported for the moderate difficulty level at an IoU threshold of 0.25. R11 and R40 denote the evaluation metric of 11 and 40 recall positions.
}
\label{tab:ood-kittiar-test}
\resizebox{0.95\textwidth}{!}{
\begin{tabular}{l|c|c|c|c|c|c|c|c|c|c|c|c|c}
\toprule
         &          & \multicolumn{3}{c|}{OoD(Mod)} & \multicolumn{3}{c|}{Pedestrian(Mod)}  &\multicolumn{3}{c|}{Car(Mod)}  &\multicolumn{3}{c}{Cyclist(Mod)}      \\
\midrule
Method  &  Recall     & $AP_{3D}$&$AP_{BEV}$ &$AP_{2D}$     & $AP_{3D}$&$AP_{BEV}$ &$AP_{2D}$ & $AP_{3D}$&$AP_{BEV}$ &$AP_{2D}$& $AP_{3D}$&$AP_{BEV}$ &$AP_{2D}$      \\ 
\midrule
OV-mono3D\cite{ovmono3d_arx24}  &R11  & 0.64&1.85&90.91& 23.75&24.14&100.0& 55.38&56.12&100.0&20.03&21.16&100.0\\ 
S3AD\cite{S3AD_mu25} &R11 &\textbf{74.35}&76.39&90.06 &48.26&49.07&64.80 & 80.20&80.41&88.54 &25.26&25.25&\textbf{37.12}\\
DDStereo(Ours) &R11 &74.23&\textbf{83.01}&\textbf{98.36} & \textbf{51.86}&\textbf{52.16}&60.33  & \textbf{87.23}&\textbf{87.31}&\textbf{89.85 }& \textbf{32.82}&\textbf{32.82}&33.39 \\ 
\textit{Improvement} & R11 & \color{red}{-0.12} & \color{blue}{+6.62} & \color{blue}{+8.30} & \color{blue}{+3.60} & \color{blue}{+3.09} & \color{red}{-4.47} & \color{blue}{+7.03} & \color{blue}{+6.90} & \color{blue}{+1.31} & \color{blue}{+7.56} & \color{blue}{+7.57} & \color{red}{-3.73} \\
\midrule
OV-mono3D\cite{ovmono3d_arx24}&R40 & 0.53&1.02&97.50& 18.84&20.01&\textbf{100.0}& 56.54&57.50&\textbf{100.0}&17.95&19.60&\textbf{100.0}\\ 
S3AD\cite{S3AD_mu25} &R40  &73.69&76.76&95.53&45.74&46.39&63.96&80.25&82.23&89.01&20.82&22.02&\textbf{34.40}\\
DDStereo(Ours) &R40 &\textbf{78.09}&\textbf{82.44}&\textbf{99.14}&\textbf{52.41}&\textbf{52.89}&62.39&\textbf{87.47}&\textbf{87.68}&93.40&\textbf{31.68}&\textbf{31.69}&32.75\\ 
\textit{Improvement} & R40 & \color{blue}{+4.40} & \color{blue}{+5.68} & \color{blue}{+3.61} & \color{blue}{+6.67} & \color{blue}{+6.50} & \color{red}{-1.57} & \color{blue}{+7.22} & \color{blue}{+5.45} & \color{blue}{+4.39} & \color{blue}{+10.86} & \color{blue}{+9.67} & \color{red}{-1.65} \\
\bottomrule
\end{tabular}}
\end{table}

We evaluate our method under the open-set setting on the AR-OoD\cite{S3AD_mu25}, and report the results in Table~\ref{tab:ood-kittiar-test}. Following the S3AD\cite{S3AD_mu25}. we focus on the moderate difficulty level and set the confidence threshold to 0.25. We report results under both R11 and R40 settings. DDStereo achieves consistently higher accuracy for both known and unknown object categories. On the unknown OoD class, DDStereo attains 74.23\% $AP_{3D}$, 83.01\% $AP_{BEV}$, 98.36\% $AP_{2D}$ under R11 and 78.09\%, 82.44\%, 99.14\% under R40, surpassing S3AD\cite{S3AD_mu25} especially in BEV localization. Moreover, DDStereo also achieves stronger performance on the known classes Pedestrian, Car, and Cyclist, demonstrating its balanced capability in both open-set and close-set 3D detection. Fig.~\ref{fig:result} presents selected qualitative results. The rhino and the bench shown in the figure belong to categories not present in the training set, but DDStereo successfully estimates their 3D positions and dimensions. For OV-Mono3D\cite{ovmono3d_arx24}, we provided ground truth text prompts during testing like "Dog". Benefiting from the broad 2D open-vocabulary coverage, it achieved 100\% 2D detection accuracy (IoU $>$ 0.25) on cars, pedestrians , and cyclists. However, its 3D detection performance is inferior to our method.

\begin{table}[t]
\centering
\small
\caption{3D Object Detection of Car Category on KITTI. (IoU$>$0.7)}
\label{kitti_test_server_3D_bev_car}
\resizebox{0.9\textwidth}{!}{%
\begin{tabular}{@{}l|c|c|ccc|ccc|c@{}}
\toprule
& & &\multicolumn{3}{c|}{$AP_{3D}(Test)$} & \multicolumn{3}{c|}{ $AP_{BEV}(Test)$} & Time \\
Method & Input&OpenSet& Easy  & Mod   & Hard  & Easy  & Mod   & Hard   & (ms)\\ \midrule
MonoFlex~\cite{MonoFLex_cvpr21} &Monocular&$\times$& 19.94 & 13.89 & 12.07 & 28.23 & 19.75 & 16.89 &30 \\

MonoJSG\cite{Monojsg_cvpr22} &Monocular&$\times$  & 24.69 & 16.14 & 13.64 & 32.59 & 21.26 & 18.18& 42 \\

GUPNet++\cite{GPUnet++_PAMI24} &Monocular &$\times$
                               & 24.99 & 16.48 & 14.58 & - & - & - &34 \\   

MonoDGP\cite{Monodgp_cvpr25}  &Monocular&$\times$
                               & 26.35 & 18.72 & 15.97 & 35.24 & 25.23 & 22.02 & 22\\ 
MonoDETR\cite{monodetr_iccv23} &Monocular &$\times$ & 25.00 &16.47 &13.58 & 33.60&  22.11&  18.60   &20 \\
                               \midrule
OC-Stereo\cite{OC-Stereo-ICRA20} &Stereo &$\times$ & 55.15&	37.60 &	30.25 &68.89 &	51.47 &	42.97 & 350 \\

SIDE\cite{SIDE_WACV22}    &Stereo &$\times$    & 47.69 & 30.82 &25.68 & - & - & -&260 \\

StereoRCNN\cite{Stereo-R-CNN-2019} &Stereo &$\times$
                               & 47.58 & 30.23 & 23.72 & 61.92 & 41.31 & 33.42  &  200 \\

DSC3D\cite{DSC3D_TCSVT25}  &Stereo   &$\times$  & \textbf{66.46 }& 42.54 & 34.04 & 74.56 & 51.21 & 42.07  & 110 \\

TLNet\cite{TLNet_CVPR19}   &Stereo  &$\times$   & 7.46  & 4.37  & 3.74  & 13.71 & 7.69  & 6.73  &  100 \\
FGAS RCNN\cite{FGAS_AEI23} &Stereo &$\times$ & 58.02 & 38.68 & 32.53 & 72.56 & 58.31 & 46.24 &    100 \\

TS3D\cite{TS3D_TITS24}   &Stereo  &$\times$    & 64.61 & 41.29 & 30.68 & 73.34 & 48.59 & 36.98  & 88 \\

YoloStereo3D\cite{YOLOStereo3D_2021} &Stereo  &$\times$
                               & 65.68 & 41.25 & 30.42 & 76.10& 50.28 & 36.86 & 80 \\ 

RT3DStereo\cite{RT3DStereo_2019}&Stereo &$\times$
                               & 29.90 & 23.28 & 18.96 & 58.81 & 46.82 & 38.38 &79 \\
ESGN\cite{ESGN_TCSVT22} &Stereo &$\times$ &65.80 &\textbf{46.39}&\textbf{38.42}&\textbf{78.10}&58.12&\textbf{49.28}& 62\\
RT3D-GMP\cite{RT3D-GMP-ITSC20} &Stereo &$\times$ & 45.79 & 38.76 & 30.00 & 69.14 & \textbf{59.00} & 45.49 &\textbf{60}\\
\midrule
\multicolumn{10}{l}{\textit{\textbf{Real-time Stereo Methods}}} \\
StereoCenterNet\cite{StereoCenter_NC22} &Stereo &$\times$
                               & 49.94 & 31.30 & 25.62 & 62.97 & 42.12 & 35.37& 43\\ 

RTS3D\cite{RTD3D_AAAI2021}    &Stereo &$\times$ & 58.51 & 37.38 & 31.12 & 72.17 & 51.79 & 43.19 &39\\ 
StereoDETR\cite{StereoDETR} &Stereo &$\times$ & 59.45& 41.17&35.13 &72.77 &\textbf{54.53} &\textbf{46.41}&\textbf{18}\\

DDStereo(Ours)        & Stereo &\checkmark   &\textbf{62.44} &	\textbf{43.97} &\textbf{36.16}  &\textbf{73.63} &	53.60 &	45.02&  24\\
\midrule
 \textit{Improvement} &  & &+2.99 & +2.80 & +1.03 & +0.86 & -0.93 & -1.39 &+6 \\
 \bottomrule
\end{tabular}}
\end{table}

\subsection{Close-Set Results on KITTI Benchmark}

We report the 3D object detection performance on the KITTI benchmark in Table~\ref{kitti_test_server_3D_bev_car}, including both the official \textit{test server} results (for Car) and offline \textit{validation set} evaluations. 
Our DDStereo achieves state-of-the-art results on the \textit{Moderate} and \textit{Hard} difficulty levels of $AP_{3D}$ with 43.97\% and 36.16\% respectively, outperforming all existing stereo-based approaches, including DSC3D~\cite{DSC3D_TCSVT25} and YOLOStereo3D~\cite{YOLOStereo3D_2021}.
Since the training set contains only 3K stereo pairs, it is insufficient for effectively training a Transformer-based architecture. Consequently, the proposed method does not surpass the YOLOStereo3D-based approaches\cite{YOLOStereo3D_2021,DSC3D_TCSVT25,TS3D_TITS24} on the Easy subset. However, it still maintains an accuracy advantage over existing real-time methods\cite{RTD3D_AAAI2021,StereoCenter_NC22}. Moreover, the Easy, Mod, and Hard splits are defined by difficulty thresholds, where the Hard set includes the Easy and Mod, thus better reflecting the overall average accuracy. In terms of efficiency, DDStereo runs at 24\,ms per frame significantly outpacing TS3D (88\,ms)~\cite{TS3D_TITS24}, RTS3D (39\,ms)~\cite{RTD3D_AAAI2021}, and YOLOStereo3D (80\,ms)\cite{YOLOStereo3D_2021}, while maintaining superior accuracy. Relative to the closed-set baseline StereoDETR\cite{StereoDETR}, DDstereo extends open-set detection capability with only 6\,ms extra latency, while simultaneously achieving better closed-set performance.

\begin{table}
\centering
\caption{3D Detection of Pedestrian and Cyclist on The KITTI Test Server.}
\label{test_on_pedestrian_mini}
\resizebox{\textwidth}{!}{
\begin{tabular}{l|c |c |c |c |c |c |c |c |c |c |c |c}
\toprule
 & \multicolumn{6}{c|}{Pedestrian} &\multicolumn{6}{c}{Cyclist} \\ 
\midrule
 & \multicolumn{2}{c|}{Easy}  & \multicolumn{2}{c|}{Mod}  & \multicolumn{2}{c|}{Hard}  & \multicolumn{2}{c|}{Easy}  & \multicolumn{2}{c|}{Mod}  & \multicolumn{2}{c}{Hard}\\ 
 \midrule
Method   & $AP_{3D}$  & $AP_{BEV}$   & $AP_{3D}$  & $AP_{BEV}$ & $AP_{3D}$  & $AP_{BEV}$ & $AP_{3D}$  & $AP_{BEV}$ & $AP_{3D}$  & $AP_{BEV}$ & $AP_{3D}$  & $AP_{BEV}$\\ 
\midrule
MonoFlex\cite{MonoFLex_cvpr21} & 9.43&10.36  & 6.31&7.36 & 5.26&6.29 & 4.17&4.41  & 2.35&2.35 & 2.04&2.50\\ 
DEVIANT\cite{Deviant_eccv22} & 13.43&14.49  & 8.65&9.77 & 7.69&8.28& 5.05&6.42  & 3.13&3.97 & 2.59&3.51 \\ 

MonoDTR\cite{Monodtr_cvpr22} &15.33&16.66 & 10.18&10.59 & 8.61&9.00&5.05&5.84 & 3.27&4.11 & 3.19&3.48 \\ 
GUPNet++\cite{GPUnet++_PAMI24}    &12.45& ---& 8.13& ---& 6.91& --- &6.71& ---& 3.91& ---& 3.80& ---\\ 
\midrule
RT3DStereo\cite{RT3DStereo_2019} & 3.28&4.72 & 2.45&3.65 &2.35&3.00&---&---&--- &---&---&---\\
ESGN\cite{ESGN_TCSVT22} &14.05&17.94 &10.27&13.03 & 9.02&11.54&13.84 &15.78&7.69&9.02&6.75&7.96 \\
RT3D-GMP\cite{RT3D-GMP-ITSC20}&16.23&19.92 & 11.41&14.22& 10.12&12.83& 18.31&20.59 & 12.99&13.92  &  10.63&12.74\\

OC-Stereo\cite{OC-Stereo-ICRA20} & 24.48&29.79& 17.58&20.80 & 15.60&18.62& 29.40&32.47&16.63&19.23  & 14.72&17.11\\ 
YoloStereo3D\cite{YOLOStereo3D_2021}  &28.49&31.01 & 19.75&20.76 & 16.48&18.41 &---&---&---&---&---&---\\ 
DSC3D\cite{DSC3D_TCSVT25}  & \underline{29.60}&\underline{32.25}& \underline{20.43}&\underline{22.57} & \underline{17.92}&\underline{19.01} &---&---&---&---&---&---\\ 
\midrule
DDStereo(Ours)  & \textbf{32.14}&\textbf{35.80} &\textbf{21.92}&\textbf{24.72} & \textbf{19.43}&\textbf{22.06}&\textbf{34.59}&\textbf{37.92}  & \textbf{21.80}&\textbf{24.36} & \textbf{18.59}&\textbf{20.87} \\ 
\bottomrule
\end{tabular}}
\end{table}

As shown in Table~\ref{test_on_pedestrian_mini}, our method DDStereo achieves the best performance across all difficulty levels of the Pedestrian class, surpassing all previous stereo methods. DDStereo achieves consistent improvements over the previous best-performing stereo method DSC3D\cite{DSC3D_TCSVT25} across all difficulty levels. 
For the Cyclist category, which poses greater challenges, only a few existing works have reported results on the KITTI test server. Among the published stereo-based methods, OC-Stereo\cite{OC-Stereo-ICRA20} achieves the strongest 3D detection performance. Compared to it, our DDStereo achieves consistent improvements across all difficulty levels. Other methods listed in Table~\ref{kitti_test_server_3D_bev_car} did not report results or release code for the Pedestrian and Cyclist classes, and thus are not included in the comparison.

On the more challenging Argoverse\cite{Argoverse_2019_CVPR} dataset, we follow the same evaluation metrics as StereoDistill\cite{StereoDistill} and present BEV and 3D detection results for the moderate Car and Pedestrian categories at 3D IoU thresholds of 0.5 and 0.25. As shown in Table \ref{tab:performance_comparison_argoverse}, our method demonstrates a clear accuracy advantage over the pure stereo approach YOLOStereo3D\cite{YOLOStereo3D_2021}, although it remains inferior to LiDAR-distillation-based methods.

\begin{table}[h]
\centering
\caption{Close-set performance comparison on Argoverse\cite{Argoverse_2019_CVPR}. }
\label{tab:performance_comparison_argoverse}
\resizebox{0.7\textwidth}{!}{%
\begin{tabular}{l|c|cc|cc}
\toprule
Method & Lidar Distill&$AP^{Car}_{3D}$& $AP^{Car}_{BEV}$ & $AP^{Ped}_{3D}$ & $AP^{Ped}_{BEV}$ \\
\midrule
DSGN\cite{DSGN} & $\checkmark$&33.68 & 42.76 & 8.58 & 8.95 \\
LIGA\cite{LIGA-Stereo_ICCV} & $\checkmark$& 34.37 & 45.47 & 10.76 & 11.01 \\
StereoDistill\cite{StereoDistill} & $\checkmark$& 37.55 & 46.99 & 13.70 & 14.04 \\
\midrule
YOLOStereo3D\cite{YOLOStereo3D_2021} & $\times$&16.65&23.87&3.65&3.94 \\
DDStereo (Ours)&$\times$&\textbf{25.74} &\textbf{30.87} & \textbf{8.52}& \textbf{8.78} \\
\textit{Improvement} &  $\times$ & \color{blue}{+9.09} & \color{blue}{+7.00} & \color{blue}{+4.87} & \color{blue}{+4.84} \\
\bottomrule
\end{tabular}}
\end{table}

\subsection{Ablation Study}

\noindent \textbf{Shared Queries.}
The dual-branch architecture separately predicts binary foreground and multi-class 3D attributes.
In the CNN-based framework, S3AD\cite{S3AD_mu25} aligns anchors in the decoding stage to compute anomaly scores.
In this work, a Transformer-based end-to-end framework is adopted, where shared queries are used for query-level alignment.
As shown in Tables~\ref{tab:Shared-Queries} and~\ref{tab:ablation-share-q}, shared queries consistently boost both closed-set and open-set performance. 
Especially for open-set detection, employing an independent query design leads to a decrease of approximately 27\% in $AP^{OoD}_{3D}$, primarily because decoding order misalignment interferes with confidence estimation.

\begin{table}[h]
  \centering
  \footnotesize
  \setlength{\tabcolsep}{1.5pt}
  
  \begin{minipage}[t]{0.5\textwidth}
    \centering
    \caption{Queries Ablation close-set KITTI\cite{KITTI} (IoU $>$ 0.7/0.5).}
    \label{tab:Shared-Queries}
    \resizebox{0.8\textwidth}{!}{
    \begin{tabular}{l|c|c|c|c|c|c}
      \toprule
       & \multicolumn{3}{c|}{$AP^{Car}_{3D}$ (0.7)} & \multicolumn{3}{c}{$AP^{Cyc}_{3D}$ (0.5)} \\
      \midrule
      Queries & Easy & Mod& Hard& Easy&Mod&Hard \\
      \midrule
      Indep. & 61.9&44.3&37.6 & 34.9&19.0&17.2 \\
      Shared & \textbf{66.9}&\textbf{48.2}&\textbf{41.1} & \textbf{37.8}&\textbf{20.4}&\textbf{18.9} \\
      \bottomrule
    \end{tabular}}
  \end{minipage}
  \hfill
  \begin{minipage}[t]{0.45\textwidth}
    \centering
    \caption{Queries Ablation on open-set AR-OoD\cite{S3AD_mu25} (MoD, IoU $>$ 0.25).}
    \label{tab:ablation-share-q}
    \resizebox{\textwidth}{!}{
    \begin{tabular}{l|c|ccc}
      \toprule
      Queries  &$AP^{OoD}_{3D}$ & $AP^{Car}_{3D}$ & $AP^{Ped}_{3D}$ & $AP^{Cyc}_{3D}$ \\
      \midrule
      Indep.  &51.19 & 72.83 & 41.20 & 26.01 \\
      Shared & \textbf{78.09} & \textbf{87.43} & \textbf{52.51} & \textbf{31.69} \\
      \bottomrule
    \end{tabular}}
  \end{minipage}
\end{table}

\noindent \textbf{Decoder Design and Depth Prediction Strategy.} We conduct an ablation study to investigate the effectiveness of the dual-decoder structure and the depth prediction strategy, as shown in Table~\ref{tab:ablation-depth_prediction}. The table also reports the performance on OoD and normal categories at a threshold of 0.25. Replacing the dual-decoder decoder with a single design brings a significant drop of 11.72\% $AP^{OoD}_{3D}$, demonstrating the benefit of decoupling foreground detection and 3D attribute estimation. For depth prediction, we adopt the average strategy from MonoDETR~\cite{monodetr_iccv23}, which computes the mean over multiple depth predictions.
When replacing the Grid Sample operation with the Average strategy in MonoDETR~\cite{monodetr_iccv23}, the $AP^{OoD}_{3D}$ decreases by 8.70\%.

\begin{table}[h]
  \centering
  \begin{minipage}[t]{0.52\textwidth}
    \centering
    \caption{Ablation of decoder and depth strategy (Mod, IoU$>$0.25).}
    \label{tab:ablation-depth_prediction}
    \resizebox{\linewidth}{!}{
      \begin{tabular}{l|c|ccc}
        \toprule
        Setting & $AP^{OoD}_{3D}$ & $AP^{Car}_{3D}$ & $AP^{Ped}_{3D}$ & $AP^{Cyc}_{3D}$ \\
        \midrule
        DDStereo (Ours) & \textbf{78.09} & \textbf{87.43} & \textbf{52.51} & \textbf{31.69} \\ 
        \midrule
        w/o Dual Decoder & 66.37 & 85.21 & 49.70 & 27.10 \\
        w/o Grid Sample  & 69.39 & 86.31 & 50.66 & 30.34 \\ 
        \bottomrule
      \end{tabular}
    }
  \end{minipage}
  \hfill 
  \begin{minipage}[t]{0.44\textwidth}
    \centering
    \caption{OoD scoring strategy comparison (Mod, IoU$>$0.25).}
    \label{tab:ood-score}
    \resizebox{\linewidth}{!}{%
      \begin{tabular}{l|ccc}
        \toprule
        Method & $AP^{OoD}_{3D}$ & $AP^{OoD}_{BEV}$ & $AP^{OoD}_{2D}$ \\
        \midrule
        MaxLogit \cite{MaxLogit_ICML22} & 67.71 & 71.52 & 84.51 \\
        MSP \cite{MSP_ICLR17}          & 75.16 & 79.28 & 92.22 \\
        MNPF (Ours)                    & \textbf{78.09} & \textbf{82.44} & \textbf{99.14} \\
        \bottomrule
      \end{tabular}
    }
  \end{minipage}
\end{table}

\noindent \textbf{OoD Score.} 
As shown in Table~\ref{tab:ood-score}, we compare different scoring strategies for open-set detection. 
Compared to MaxLogit\cite{MaxLogit_ICML22}, which does not use normalization, and MSP\cite{MSP_ICLR17}, which utilizes Softmax normalization, the Sigmoid normalization used DDStereo achieves higher performance. This could be attributed to the use of Sigmoid-based Focal loss classification loss during the training phase. Our strategy of subtracting the highest known-class score from the binary foreground score is simple yet effective.

\noindent \textbf{Generalization with Limited Extra Data.} 
S3AD\cite{S3AD_mu25} proposed using synthetic datasets to generalize scale prediction capabilities, with its open-set detection performance improving as more supplementary data is introduced. In Table~\ref{tab:ar-data}, we compare the performance of DDStereo and S3AD under limited additional data. When only 10\% (400 images) of the extra synthetic dataset is added, our method demonstrates a more pronounced performance advantage.

\begin{table}[h]
  \centering
  \begin{minipage}[t]{0.41\textwidth}
    \centering    \caption{Generalization with Limited Extra Data.}
    \label{tab:ar-data}
    \resizebox{\linewidth}{!}{%
      \begin{tabular}{l|c|ccc}
        \toprule
        Method & AR-ExD & $AP^{OoD}_{3D}$ & $AP^{Ped}_{3D}$ & $AP^{Cyc}_{3D}$ \\
        \midrule
        S3AD \cite{S3AD_mu25} & $\times 10\%$ & 24.45 & 42.88 & 21.58 \\ 
        DDStereo & $\times 10\%$ & \textbf{54.65} & \textbf{47.76} & \textbf{29.27} \\ 
        \midrule
        S3AD \cite{S3AD_mu25} & $\times 100\%$ & 73.69 & 45.74 & 20.82 \\ 
        DDStereo & $\times 100\%$ & \textbf{78.09} & \textbf{52.51} & \textbf{31.69} \\
        \bottomrule
      \end{tabular}
    }
  \end{minipage}
  \hfill 
  \begin{minipage}[t]{0.55\textwidth}
    \centering
    \caption{Comparison of computational load. Values in parentheses is our tested speeds.
    }
    \label{Gflops_params}
    \resizebox{0.9\linewidth}{!}{%
      \begin{tabular}{l|c|c|c|c}
        \toprule
        Method & OoD & Time(ms) & GFlops & Params \\
        \midrule

        YOLOStereo3D \cite{YOLOStereo3D_2021} & $\times$ & 80 (71*) & 177.82 & 107.6 M \\ 
        StereoDETR \cite{StereoDETR} & $\times$ & 18 (18*) & 59.80 & 10.2 M \\ 
        MonoDETR \cite{monodetr_iccv23} & $\times$ & 20 (19*) & 62.96 & 37.7 M \\
        MonoDGP \cite{Monodgp_cvpr25}   & $\times$ & 22 (20*) & 71.79 & 42.2 M \\
        \midrule
        S3AD \cite{S3AD_mu25} & $\checkmark$ & 75 (75*) & 180.71 & 109.5 M\\ 
        \textbf{DDStereo (Ours)} & $\checkmark$ & 24 & 62.65 & 19.6 M \\
        \bottomrule
      \end{tabular}
    }
  \end{minipage}
\end{table}

\subsection{Model Efficiency}

We compare the computational efficiency and parameter size of our model with both monocular and stereo-based baselines in Table~\ref{Gflops_params}. In terms of computational cost, DDStereo requires only 62.65 GFLOPs, lower than the open-set stereo method S3AD\cite{S3AD_mu25} and even slightly lower than MonoDETR\cite{monodetr_iccv23} and MonoDGP\cite{Monodgp_cvpr25}. Furthermore, our model contains only 19.6\,M parameters, which is nearly half the size of the monocular models, and less than one fifth of S3AD\cite{S3AD_mu25}. These results demonstrate that our dual-decoder stereo architecture achieves a favorable balance between accuracy and efficiency, making it suitable for real-time deployment in autonomous driving systems. Compared to StereoDETR\cite{StereoDETR}, the fastest existing closed-set stereo detector, DDStereo expands its capability to arbitrary OoD obstacle detection with a marginal overhead of only 6\,ms in latency and 2.75 GFLOPs.

\subsection{Visualization}

\begin{figure}[t]
\centering
\includegraphics[height=6cm]{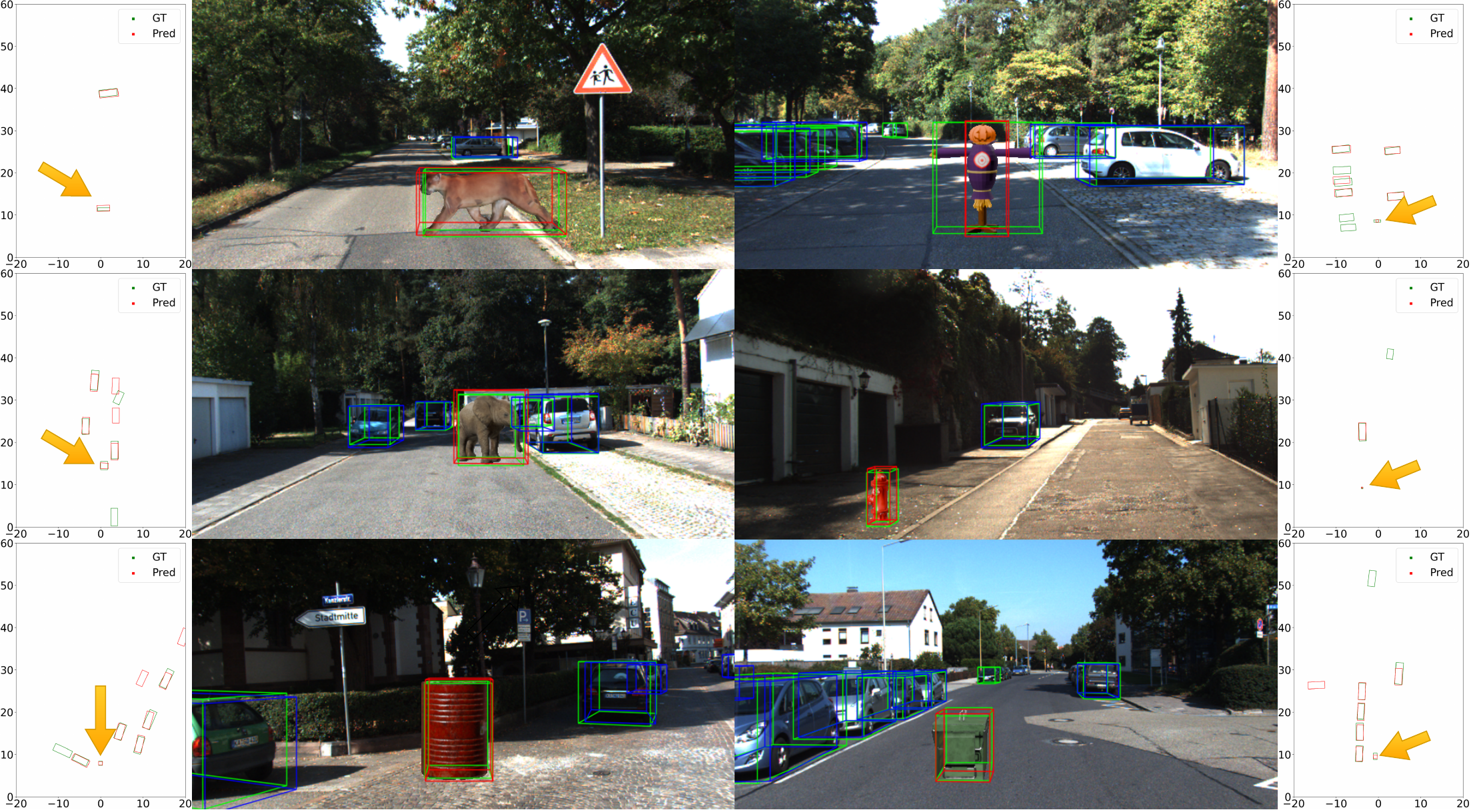}
\caption{Visualization results with 3D boxes and BEV map. In the RGB images, {\color{red}red} 3D boxes represent predicted anomaly categories, {\color{blue}blue} 3D boxes represent predicted known categories, and {\color{green}green} boxes indicate the ground truth.
}
\label{fig:result}
\end{figure}

\noindent \textbf{Image-editing-based synthetic datasets.} Fig. \ref{fig:result} visualizes the detection of anomaly classes on the AR-OoD\cite{S3AD_mu25} dataset. Despite these categories and similar ones being entirely absent during training, our approach leverages stereo disparity to accurately estimate the BEV layout of novel classes, though some scale estimation errors persist.

\begin{figure}[h]
\centering
\includegraphics[height=5cm]{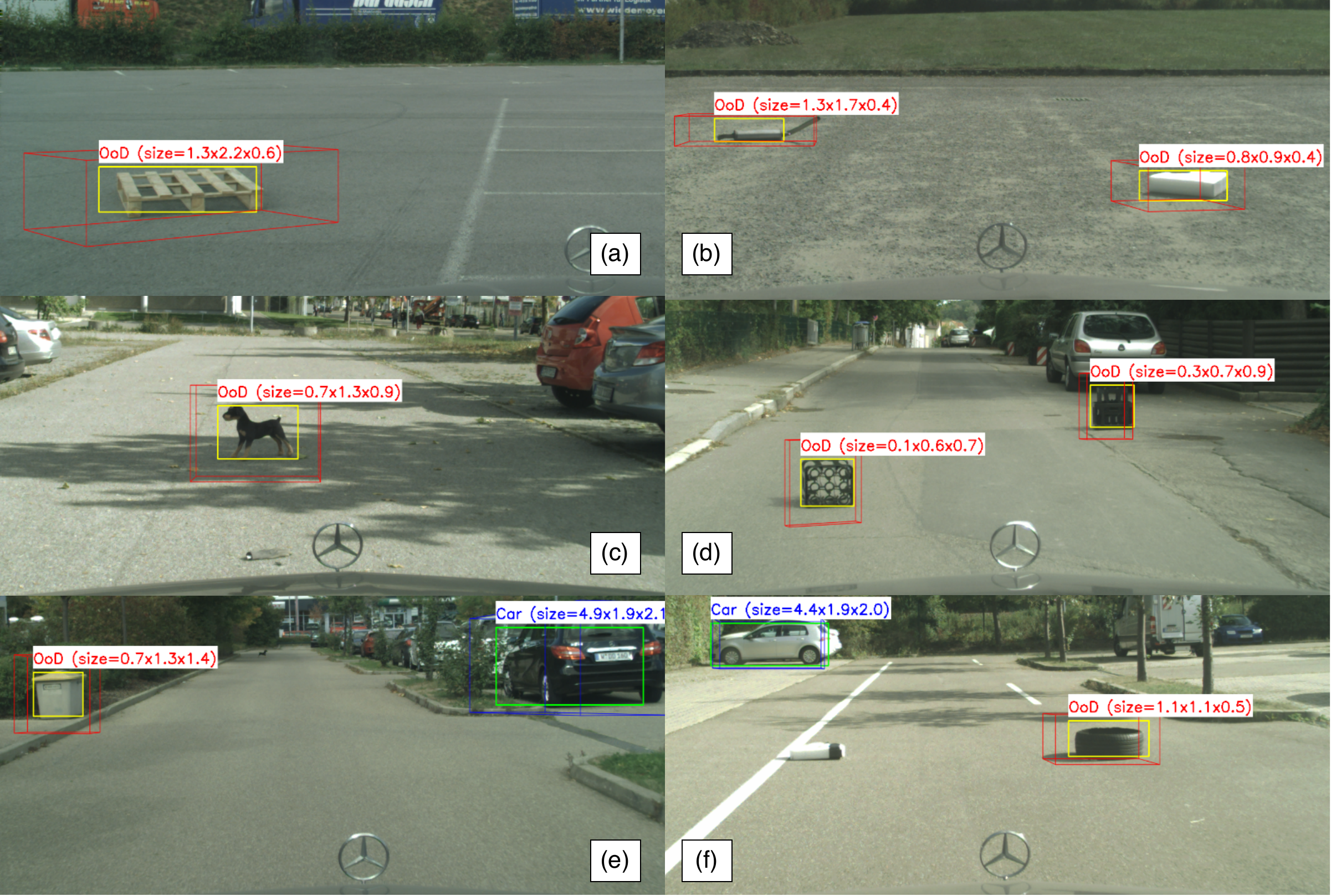}
\caption{Zero-shot test on LaF\cite{LOF}. {\color{yellow}yellow}: 2D predictions; {\color{red}red}: 3D projections.}
\label{fig:test_on_laf}
\end{figure}

\noindent \textbf{In-the-Wild Validation.}
To further evaluate real-world performance, we perform cross-camera testing on LaF\cite{LOF}, a road anomaly segmentation dataset providing paired stereo images. Our model is trained solely on 7.7K samples from AR-OoD and KITTI. As visualized in Fig.~\ref{fig:test_on_laf}, despite misaligned camera parameters, the disparity-based method successfully detects unknown obstacles. Yellow and red boxes represent predicted 2D and 3D anomaly bounding boxes, respectively. Subfigures (e) and (f) present failure cases: a false positive detection of a roadside trash can and a missed white obstacle, respectively. Overall, with merely 7K training samples, our model demonstrates reliable generalization for stereo 3D anomaly detection.

\section{Conclusion and Limitations}
We propose DDStereo, a dual-decoder transformer that unifies open-set detection with real-time stereo 3D detection. By leveraging shared queries and a streamlined architecture, DDStereo achieves a superior accuracy-efficiency trade-off, marking the first stereo anomaly detector to match monocular close-set baselines in speed. However, limitations remain: the model’s scalability across larger, more diverse datasets requires further validation, and its generalization to varied camera parameters is constrained by geometric sensitivities. Addressing these via robust rectification and large-scale pre-training will be the focus of future work.

\section*{Acknowledgements}
This work was supported in part by the 6G Science and Technology Innovation and Future Industry Cultivation Special Project of Shanghai Municipal Science and Technology Commission under Grant 24DP1501001, in part by the National High Quality Program under Grant TC220H07D, and in part by the Xi'an Jiaotong-Liverpool University under Grant for ILAI.

%
%
\bibliographystyle{splncs04}
\bibliography{main}

@String(CVPR  = {IEEE Conf. Comput. Vis. Pattern Recog.})

@String(ICCV  = {Int. Conf. Comput. Vis.})

@String(AAAI  = {AAAI})

@String(CVPR  = {CVPR})

@String(ICCV  = {ICCV})

@inproceedings{ovmono3d_arx24,
  author={Yao, Jin and Gu, Hao and Chen, Xuweiyi and Wang, Jiayun and Cheng, Zezhou},
  booktitle={2026 International Conference on 3D Vision (3DV)}, 
  title={Open Vocabulary Monocular 3D Object Detection}, 
  year={2026},
  volume={},
  number={},
  pages={1801-1811}}

@InProceedings{DetAny3D_iccv25,
    author    = {Zhang, Hanxue and Jiang, Haoran and Yao, Qingsong and Sun, Yanan and Zhang, Renrui and Zhao, Hao and Li, Hongyang and Zhu, Hongzi and Yang, Zetong},
    title     = {Detect Anything 3D in the Wild},
    booktitle = {Proceedings of the IEEE/CVF International Conference on Computer Vision (ICCV)},
    month     = {October},
    year      = {2025},
    pages     = {5048-5059}
}

@inproceedings{
gu2021open,
title={Open-vocabulary Object Detection via Vision and Language Knowledge Distillation},
author={Xiuye Gu and Tsung-Yi Lin and Weicheng Kuo and Yin Cui},
booktitle={International Conference on Learning Representations},
year={2022}
}

@inproceedings{wu2023cora,
    author    = {Wu, Xiaoshi and Zhu, Feng and Zhao, Rui and Li, Hongsheng},
    title     = {CORA: Adapting CLIP for Open-Vocabulary Detection With Region Prompting and Anchor Pre-Matching},
    booktitle = {Proceedings of the IEEE/CVF Conference on Computer Vision and Pattern Recognition (CVPR)},
    month     = {June},
    year      = {2023},
    pages     = {7031-7040}
}

@inproceedings{chuang2024generative,
  author    = {Lin, Chuang and Jiang, Yi and Qu, Lizhen and Yuan, Zehuan and Cai, Jianfei},
    title     = {Generative Region-Language Pretraining for Open-Ended Object Detection},
    booktitle = {Proceedings of the IEEE/CVF Conference on Computer Vision and Pattern Recognition (CVPR)},
    month     = {June},
    year      = {2024},
    pages     = {13958-13968}
}

@inproceedings{yao2024detclipv3,
   author    = {Yao, Lewei and Pi, Renjie and Han, Jianhua and Liang, Xiaodan and Xu, Hang and Zhang, Wei and Li, Zhenguo and Xu, Dan},
    title     = {DetCLIPv3: Towards Versatile Generative Open-vocabulary Object Detection},
    booktitle = {Proceedings of the IEEE/CVF Conference on Computer Vision and Pattern Recognition (CVPR)},
    month     = {June},
    year      = {2024},
    pages     = {27391-27401}
}

@inproceedings{yao2023detclipv2,
      author    = {Yao, Lewei and Han, Jianhua and Liang, Xiaodan and Xu, Dan and Zhang, Wei and Li, Zhenguo and Xu, Hang},
    title     = {DetCLIPv2: Scalable Open-Vocabulary Object Detection Pre-Training via Word-Region Alignment},
    booktitle = {Proceedings of the IEEE/CVF Conference on Computer Vision and Pattern Recognition (CVPR)},
    month     = {June},
    year      = {2023},
    pages     = {23497-23506}
}

@inproceedings{cheng2024yolo,
  title={Yolo-world: Real-time open-vocabulary object detection},
  author={Cheng, Tianheng and Song, Lin and Ge, Yixiao and Liu, Wenyu and Wang, Xinggang and Shan, Ying},
  booktitle={Proceedings of the IEEE/CVF Conference on Computer Vision and Pattern Recognition},
  pages={16901--16911},
  year={2024}
}

@inproceedings{liu2023grounding,
  title={Grounding dino: Marrying dino with grounded pre-training for open-set object detection},
  author={Liu, Shilong and Zeng, Zhaoyang and Ren, Tianhe and Li, Feng and Zhang, Hao and Yang, Jie and Jiang, Qing and Li, Chunyuan and Yang, Jianwei and Su, Hang and others},
  booktitle={European conference on computer vision},
  pages={38--55},
  year={2024},
  organization={Springer}
}

@inproceedings{cen2021open,
  author={Cen, Jun and Yun, Peng and Cai, Junhao and Wang, Michael Yu and Liu, Ming},
  booktitle={2021 International Conference on 3D Vision (3DV)}, 
  title={Open-set 3D Object Detection}, 
  year={2021},
  volume={},
  number={},
  pages={869-878}}

@inproceedings{lu2023open,
  title={Open-vocabulary point-cloud object detection without 3d annotation},
  author={Lu, Yuheng and Xu, Chenfeng and Wei, Xiaobao and Xie, Xiaodong and Tomizuka, Masayoshi and Keutzer, Kurt and Zhang, Shanghang},
  booktitle={Proceedings of the IEEE/CVF Conference on Computer Vision and Pattern Recognition (CVPR)},
  pages={1190--1199},
  month     = {June},
  year={2023}
}

@inproceedings{xue2023ulip,
    author    = {Xue, Le and Gao, Mingfei and Xing, Chen and Mart{\'\i}n-Mart{\'\i}n, Roberto and Wu, Jiajun and Xiong, Caiming and Xu, Ran and Niebles, Juan Carlos and Savarese, Silvio},
    title     = {ULIP: Learning a Unified Representation of Language, Images, and Point Clouds for 3D Understanding},
    booktitle = {Proceedings of the IEEE/CVF Conference on Computer Vision and Pattern Recognition (CVPR)},
    month     = {June},
    year      = {2023},
    pages     = {1179-1189}

}

@inproceedings{zhang2022pointclip,
    author    = {Zhang, Renrui and Guo, Ziyu and Zhang, Wei and Li, Kunchang and Miao, Xupeng and Cui, Bin and Qiao, Yu and Gao, Peng and Li, Hongsheng},
    title     = {PointCLIP: Point Cloud Understanding by CLIP},
    booktitle = {Proceedings of the IEEE/CVF Conference on Computer Vision and Pattern Recognition (CVPR)},
    month     = {June},
    year      = {2022},
    pages     = {8552-8562}
}

@inproceedings{cao2024coda,
  author = {Cao, Yang and Yihan, Zeng and Xu, Hang and Xu, Dan},
 booktitle = {Advances in Neural Information Processing Systems},
 editor = {A. Oh and T. Naumann and A. Globerson and K. Saenko and M. Hardt and S. Levine},
 pages = {71862--71873},
 publisher = {Curran Associates, Inc.},
 title = {CoDA: Collaborative Novel Box Discovery and Cross-modal Alignment for Open-vocabulary 3D Object Detection},
 volume = {36},
 year = {2023}
}

@inproceedings{wang2024ov,
  title={Ov-uni3detr: Towards unified open-vocabulary 3d object detection via cycle-modality propagation},
  author={Wang, Zhenyu and Li, Yali and Liu, Taichi and Zhao, Hengshuang and Wang, Shengjin},
  booktitle={European Conference on Computer Vision},
  pages={73--89},
  year={2024},
  organization={Springer}
}

@inproceedings{dhamija2020overlooked,
  title={The overlooked elephant of object detection: Open set},
  author={Dhamija, Akshay and Gunther, Manuel and Ventura, Jonathan and Boult, Terrance},
  pages={1021--1030},
  booktitle = {Proceedings of the IEEE/CVF Winter Conference on Applications of Computer Vision (WACV)},
month = {March},
year = {2020}
}

@inproceedings{han2022expanding,
 author    = {Han, Jiaming and Ren, Yuqiang and Ding, Jian and Pan, Xingjia and Yan, Ke and Xia, Gui-Song},
    title     = {Expanding Low-Density Latent Regions for Open-Set Object Detection},
    booktitle = {Proceedings of the IEEE/CVF Conference on Computer Vision and Pattern Recognition (CVPR)},
    month     = {June},
    year      = {2022},
    pages     = {9591-9600}}

@inproceedings{gupta2022ow,
    author    = {Gupta, Akshita and Narayan, Sanath and Joseph, K J and Khan, Salman and Khan, Fahad Shahbaz and Shah, Mubarak},
    title     = {OW-DETR: Open-World Detection Transformer},
    booktitle = {Proceedings of the IEEE/CVF Conference on Computer Vision and Pattern Recognition (CVPR)},
    month     = {June},
    year      = {2022},
    pages     = {9235-9244}
}

@inproceedings{joseph2021towards,
  author    = {Joseph, K J and Khan, Salman and Khan, Fahad Shahbaz and Balasubramanian, Vineeth N},
    title     = {Towards Open World Object Detection},
    booktitle = {Proceedings of the IEEE/CVF Conference on Computer Vision and Pattern Recognition (CVPR)},
    month     = {June},
    year      = {2021},
    pages     = {5830-5840}
}

@article{yao2022detclip,
  title={Detclip: Dictionary-enriched visual-concept paralleled pre-training for open-world detection},
  author={Yao, Lewei and Han, Jianhua and Wen, Youpeng and Liang, Xiaodan and Xu, Dan and Zhang, Wei and Li, Zhenguo and Xu, Chunjing and Xu, Hang},
  journal={Advances in Neural Information Processing Systems},
  volume={35},
  pages={9125--9138},
  year={2022}
}

@inproceedings{20223DOS,
  author = {Alliegro, Antonio and Cappio Borlino, Francesco and Tommasi, Tatiana},
 booktitle = {Advances in Neural Information Processing Systems},
 pages = {21228--21240},
 publisher = {Curran Associates, Inc.},
 title = {3DOS: Towards 3D Open Set Learning - Benchmarking and Understanding Semantic Novelty Detection on Point Clouds},
 volume = {35},
 year = {2022}
}

@inproceedings{MonoFLex_cvpr21,
    author    = {Zhang, Yunpeng and Lu, Jiwen and Zhou, Jie},
    title     = {Objects Are Different: Flexible Monocular 3D Object Detection},
    booktitle = {Proceedings of the IEEE/CVF Conference on Computer Vision and Pattern Recognition (CVPR)},
    month     = {June},
    year      = {2021},
    pages     = {3289-3298}
}

@inproceedings{Monojsg_cvpr22,
 author    = {Lian, Qing and Li, Peiliang and Chen, Xiaozhi},
    title     = {MonoJSG: Joint Semantic and Geometric Cost Volume for Monocular 3D Object Detection},
    booktitle = {Proceedings of the IEEE/CVF Conference on Computer Vision and Pattern Recognition (CVPR)},
    month     = {June},
    year      = {2022},
    pages     = {1070-1079}
}

@article{GPUnet++_PAMI24,
  author={Lu, Yan and Ma, Xinzhu and Yang, Lei and Zhang, Tianzhu and Liu, Yating and Chu, Qi and He, Tong and Li, Yonghui and Ouyang, Wanli},
  journal={IEEE Transactions on Pattern Analysis and Machine Intelligence}, 
  title={GUPNet++: Geometry Uncertainty Propagation Network for Monocular 3D Object Detection}, 
  year={2025},
  volume={47},
  number={2},
  pages={900-915}}

@inproceedings{Monodgp_cvpr25,
    author    = {Pu, Fanqi and Wang, Yifan and Deng, Jiru and Yang, Wenming},
    title     = {MonoDGP: Monocular 3D Object Detection with Decoupled-Query and Geometry-Error Priors},
    booktitle = {Proceedings of the IEEE/CVF Conference on Computer Vision and Pattern Recognition (CVPR)},
    month     = {June},
    year      = {2025},
    pages     = {6520-6530}
}

@inproceedings{monodetr_iccv23,
    author    = {Zhang, Renrui and Qiu, Han and Wang, Tai and Guo, Ziyu and Cui, Ziteng and Qiao, Yu and Li, Hongsheng and Gao, Peng},
    title     = {MonoDETR: Depth-guided Transformer for Monocular 3D Object Detection},
    booktitle = {Proceedings of the IEEE/CVF International Conference on Computer Vision (ICCV)},
    month     = {October},
    year      = {2023},
    pages     = {9155-9166}
}

@INPROCEEDINGS{OC-Stereo-ICRA20,
  author={Pon, Alex D. and Ku, Jason and Li, Chengyao and Waslander, Steven L.},
  booktitle={2020 IEEE International Conference on Robotics and Automation (ICRA)}, 
  title={Object-Centric Stereo Matching for 3D Object Detection}, 
  year={2020},
  volume={},
  number={},
  pages={8383-8389}}

@inproceedings{SIDE_WACV22,
  author    = {Peng, Xidong and Zhu, Xinge and Wang, Tai and Ma, Yuexin},
    title     = {SIDE: Center-Based Stereo 3D Detector With Structure-Aware Instance Depth Estimation},
    booktitle = {Proceedings of the IEEE/CVF Winter Conference on Applications of Computer Vision (WACV)},
    month     = {January},
    year      = {2022},
    pages     = {119-128}
}

@inproceedings{Stereo-R-CNN-2019,  
 title={Stereo R-CNN Based 3D Object Detection for Autonomous Driving}, 
 booktitle={2019 IEEE/CVF Conference on Computer Vision and Pattern Recognition (CVPR)}, 
 author={Li, Peiliang and Chen, Xiaozhi and Shen, Shaojie}, 
 year={2019}, 
 month={Jun}, 
 language={en-US} 
 }

@ARTICLE{DSC3D_TCSVT25,
  author={Chen, Jiawei and Song, Qi and Guo, Wenzhong and Huang, Rui},
  journal={IEEE Transactions on Circuits and Systems for Video Technology}, 
  title={DSC3D: Deformable Sampling Constraints in Stereo 3D Object Detection for Autonomous Driving}, 
  year={2025},
  volume={35},
  number={3},
  pages={2794-2805}}

@inproceedings{TLNet_CVPR19,
author = {Qin, Zengyi and Wang, Jinglu and Lu, Yan},
title = {Triangulation Learning Network: From Monocular to Stereo 3D Object Detection},
booktitle = {Proceedings of the IEEE/CVF Conference on Computer Vision and Pattern Recognition (CVPR)},
month = {June},
pages={7615-7623},
year = {2019}
}

@article{FGAS_AEI23,
  title={An efficient 3D object detection method based on fast guided anchor stereo RCNN},
  author={Tao, Chongben and Cao, Chunlin and Cheng, Hanjing and Gao, Zhen and Luo, Xizhao and Zhang, Zuofeng and Zheng, Sifa},
  journal={Advanced Engineering Informatics},
  volume={57},
  pages={102069},
  year={2023},
  publisher={Elsevier}
}

@ARTICLE{TS3D_TITS24,
  author={Sun, Hanqing and Pang, Yanwei and Cao, Jiale and Xie, Jin and Li, Xuelong},
  journal={IEEE Transactions on Intelligent Transportation Systems}, 
  title={Transformer-Based Stereo-Aware 3D Object Detection From Binocular Images}, 
  year={2024},
  volume={25},
  number={12},
  pages={19675-19687}}

@inproceedings{YOLOStereo3D_2021,  
author={Liu, Yuxuan and Wang, Lujia and Liu, Ming},
  booktitle={2021 IEEE International Conference on Robotics and Automation (ICRA)}, 
  title={YOLOStereo3D: A Step Back to 2D for Efficient Stereo 3D Detection}, 
  year={2021},
  volume={},
  number={},
  pages={13018-13024}}

@inproceedings{RT3DStereo_2019,  
title={Realtime 3D Object Detection for Automated Driving Using Stereo Vision and Semantic Information}, 
booktitle={2019 IEEE Intelligent Transportation Systems Conference (ITSC)}, 
author={Konigshof, Hendrik and Salscheider, Niels Ole and Stiller, Christoph}, 
year={2019}, 
month={Oct}, 
pages={1405-1410},
language={en-US} 
}

@inproceedings{RT3D-GMP-ITSC20,
  title={Learning-based shape estimation with grid map patches for realtime 3D object detection for automated driving},
  author={Konigshof, Hendrik and Stiller, Christoph},
  booktitle={2020 IEEE 23rd International conference on intelligent transportation systems (ITSC)},
  pages={1--6},
  year={2020},
  organization={IEEE}
}

@article{StereoCenter_NC22,
  title={Stereo CenterNet-based 3D object detection for autonomous driving},
  author = {Yuguang Shi and Yu Guo and Zhenqiang Mi and Xinjie Li},
  journal={Neurocomputing},
  volume={471},
  pages={219--229},
  year={2022},
  publisher={Elsevier}
}

@inproceedings{RTD3D_AAAI2021,
  title={Rts3d: Real-time stereo 3d detection from 4d feature-consistency embedding space for autonomous driving},
  author={Li, Peixuan and Su, Shun and Zhao, Huaici},
  booktitle={Proceedings of the AAAI Conference on Artificial Intelligence},
  volume={35},
  pages={1930--1939},
  year={2021}
}

@article{S3AD_mu25,
author={Mu, Shiyi and Gu, Zichong and Lyu, Hanqi and Gao, Yilin and Xu, Shugong},
  journal={IEEE Internet of Things Journal}, 
  title={Stereo-based 3D Anomaly Object Detection for Autonomous Driving: A New Dataset and Baseline}, 
  year={2026},
  volume={},
  number={},
  pages={1-1}}

@inproceedings{MSP_ICLR17,
  title={A Baseline for Detecting Misclassified and Out-of-Distribution Examples in Neural Networks},
  author={Hendrycks, Dan and Gimpel, Kevin},
  booktitle={International Conference on Learning Representations},
  year={2017}
}

@InProceedings{MaxLogit_ICML22,
  title = 	 {Scaling Out-of-Distribution Detection for Real-World Settings},
  author =       {Hendrycks, Dan and Basart, Steven and Mazeika, Mantas and Zou, Andy and Kwon, Joseph and Mostajabi, Mohammadreza and Steinhardt, Jacob and Song, Dawn},
  booktitle = 	 {Proceedings of the 39th International Conference on Machine Learning},
  pages = 	 {8759--8773},
  year = 	 {2022},
  volume = 	 {162},
  month = 	 {Jul},
  publisher =    {PMLR}
}

@inproceedings{KITTI,   
title={Are we ready for autonomous driving? The KITTI vision benchmark suite},   
booktitle={2012 IEEE Conference on Computer Vision and Pattern Recognition},  
author={Geiger, A. and Lenz, P. and Urtasun, R.}, 
year={2012},  
month={Jun},  
pages={3354-3361} }

@inproceedings{Monodtr_cvpr22,
  author    = {Huang, Kuan-Chih and Wu, Tsung-Han and Su, Hung-Ting and Hsu, Winston H.},
    title     = {MonoDTR: Monocular 3D Object Detection With Depth-Aware Transformer},
    booktitle = {Proceedings of the IEEE/CVF Conference on Computer Vision and Pattern Recognition (CVPR)},
    month     = {June},
    year      = {2022},
    pages     = {4012-4021}
}

@inproceedings{LOF,
  author={Pinggera, Peter and Ramos, Sebastian and Gehrig, Stefan and Franke, Uwe and Rother, Carsten and Mester, Rudolf},
  booktitle={2016 IEEE/RSJ International Conference on Intelligent Robots and Systems (IROS)}, 
  title={Lost and Found: detecting small road hazards for self-driving vehicles}, 
  year={2016},
  volume={},
  number={},
  pages={1099-1106}}

@inproceedings{StereoDistill,
title={Stereodistill: Pick the cream from lidar for distilling stereo-based 3d object detection},
  author={Liu, Zhe and Ye, Xiaoqing and Tan, Xiao and Ding, Errui and Bai, Xiang},
  booktitle={Proceedings of the AAAI Conference on Artificial Intelligence},
  volume={37},
  pages={1790--1798},
  year={2023}
}

@inproceedings{Deviant_eccv22,
  title={Deviant: Depth equivariant network for monocular 3d object detection},
  author={Kumar, Abhinav and Brazil, Garrick and Corona, Enrique and Parchami, Armin and Liu, Xiaoming},
  booktitle={European Conference on Computer Vision},
  pages={664--683},
  year={2022},
  organization={Springer}
}

@InProceedings{LIGA-Stereo_ICCV,
    author    = {Guo, Xiaoyang and Shi, Shaoshuai and Wang, Xiaogang and Li, Hongsheng},
    title     = {LIGA-Stereo: Learning LiDAR Geometry Aware Representations for Stereo-Based 3D Detector},
    booktitle = {Proceedings of the IEEE/CVF International Conference on Computer Vision (ICCV)},
    month     = {October},
    year      = {2021},
    pages     = {3153-3163}
}

@InProceedings{DSGN,
author = {Chen, Yilun and Liu, Shu and Shen, Xiaoyong and Jia, Jiaya},
title = {DSGN: Deep Stereo Geometry Network for 3D Object Detection},
booktitle = {Proceedings of the IEEE/CVF Conference on Computer Vision and Pattern Recognition (CVPR)},
month = {June},
year = {2020}
}

@InProceedings{Waymo_2020_CVPR,
author = {Sun, Pei and Kretzschmar, Henrik and Dotiwalla, Xerxes and Chouard, Aurelien and Patnaik, Vijaysai and Tsui, Paul and Guo, James and Zhou, Yin and Chai, Yuning and Caine, Benjamin and Vasudevan, Vijay and Han, Wei and Ngiam, Jiquan and Zhao, Hang and Timofeev, Aleksei and Ettinger, Scott and Krivokon, Maxim and Gao, Amy and Joshi, Aditya and Zhang, Yu and Shlens, Jonathon and Chen, Zhifeng and Anguelov, Dragomir},
title = {Scalability in Perception for Autonomous Driving: Waymo Open Dataset},
booktitle = {Proceedings of the IEEE/CVF Conference on Computer Vision and Pattern Recognition (CVPR)},
month = {June},
year = {2020},
    pages     = {2446-2454}
}

@InProceedings{nuScenes_2020_CVPR,
author = {Caesar, Holger and Bankiti, Varun and Lang, Alex H. and Vora, Sourabh and Liong, Venice Erin and Xu, Qiang and Krishnan, Anush and Pan, Yu and Baldan, Giancarlo and Beijbom, Oscar},
title = {nuScenes: A Multimodal Dataset for Autonomous Driving},
booktitle = {Proceedings of the IEEE/CVF Conference on Computer Vision and Pattern Recognition (CVPR)},
month = {June},
year = {2020}
}

@InProceedings{Argoverse_2019_CVPR,
author = {Chang, Ming-Fang and Lambert, John and Sangkloy, Patsorn and Singh, Jagjeet and Bak, Slawomir and Hartnett, Andrew and Wang, De and Carr, Peter and Lucey, Simon and Ramanan, Deva and Hays, James},
title = {Argoverse: 3D Tracking and Forecasting With Rich Maps},
booktitle = {Proceedings of the IEEE/CVF Conference on Computer Vision and Pattern Recognition (CVPR)},
month = {June},
year = {2019}
}

@ARTICLE{ESGN_TCSVT22,
  author={Gao, Aqi and Pang, Yanwei and Nie, Jing and Shao, Zhuang and Cao, Jiale and Guo, Yishun and Li, Xuelong},
  journal={IEEE Transactions on Circuits and Systems for Video Technology}, 
  title={ESGN: Efficient Stereo Geometry Network for Fast 3D Object Detection}, 
  year={2024},
  volume={34},
  number={4},
  pages={2000-2009}}

@ARTICLE{StereoDETR,
  author={Mu, Shiyi and Gu, Zichong and Ai, Zhiqi and Liu, Anqi and Gao, Yilin and Xu, Shugong},
  journal={IEEE Transactions on Circuits and Systems for Video Technology}, 
  title={StereoDETR: Stereo-Based Transformer for 3D Object Detection}, 
  year={2026},
  volume={36},
  number={4},
  pages={4371-4383}}
\end{document}